\documentclass[10pt,twocolumn,letterpaper]{article}

\usepackage{titling}
\usepackage{cvpr}
\usepackage{times}
\usepackage{epsfig}
\usepackage{graphicx}
\usepackage{amsmath}
\usepackage{amssymb}

\usepackage{algorithm}
\usepackage[noend]{algpseudocode}
\usepackage{subcaption}

\usepackage[pagebackref=true,breaklinks=true,letterpaper=true,colorlinks,bookmarks=false]{hyperref}

\begin{document}
\cvprfinalcopy 

\def\cvprPaperID{****} 
\def\httilde{\mbox{\tt\raisebox{-.5ex}{\symbol{126}}}}

\ifcvprfinal\pagestyle{empty}\fi
\title{Are All Training Examples Created Equal? An Empirical Study}

\author{Kailas Vodrahalli\\
UC Berkeley\\
{\tt\small kailasv@berkeley.edu}
\and
Ke Li\\
UC Berkeley\\
{\tt\small ke.li@eecs.berkeley.edu}
\and
Jitendra Malik\\
UC Berkeley\\
{\tt\small malik@eecs.berkeley.edu}
}

\maketitle

\begin{abstract}
  Modern computer vision algorithms often rely on very large training datasets. However, it is conceivable that a carefully selected subsample of the dataset is sufficient for training. In this paper, we propose a gradient-based importance measure that we use to empirically analyze relative importance of training images in four datasets of varying complexity. We find that in some cases, a small subsample is indeed sufficient for training. For other datasets, however, the relative differences in importance are negligible. These results have important implications for active learning on deep networks. Additionally, our analysis method can be used as a general tool to better understand diversity of training examples in datasets.
\end{abstract}


\section{Introduction} \label{Introduction}

Deep learning has achieved remarkable success in recent years, largely made possible by more powerful hardware and an abundance of training data. In particular, computer vision has been enabled by deep convolutional neural networks to attain near or exceeding human performance on a variety of tasks including image classification \cite{he2015delving}, object segmentation \cite{chen2018deeplab}, pose recognition \cite{newell2016stacked}, image synthesis \cite{isola2017image}, and many others. These deep networks are typically trained using stochastic gradient methods where data is subsampled in minibatches and the network parameters are updated by the gradient of the parameter weights relative to a loss function for the given minibatch (modulated by some learning rate and other hyperparameters).

As these deep networks typically have millions of learnable parameters, they require millions of images for training. In many cases, large datasets are not publically available and are expensive to collect and annotate. So, a common procedure is to pretrain a network using a large image dataset and subsequently finetune the network for a specific application using a smaller, application-specific dataset.

Despite the importance of these datasets for many applications, there is a dearth of analysis on their properties. In this paper, we seek to remedy this through an analysis over several well-known datasets: MNIST \cite{lecun1998mnist}, CIFAR-10 \cite{krizhevsky2009learning}, CIFAR-100 \cite{krizhevsky2009learning}, and ImageNet \cite{deng2009imagenet}. In particular, we are interested in understanding the relative importance of images for training a deep neural network and determining how diverse these datasets are. For the purposes of this paper, we restrict ourselves to image datasets and select image classification as our application. It is, however, feasible to apply our techniques to other datasets and tasks. We conduct this analysis by computing the gradient magnitude of the loss corresponding to each individual training image at the end of training to determine a relative importance score for each image. We then retrain our network on subsets of the data selected based on our importance measure to determine how well these subsets capture the distribution of the entire dataset. 

We propose two attributes to compare datasets: simplicity and redundancy. Using these attributes, we conclude that while some common datasets like CIFAR-10/100 and ImageNet do indeed contain diverse images, others like MNIST have much redundancy in data. Our analysis has implications for the difficulty of performing active learning in the context of deep learning and for dataset collection practices.


Our paper is organized as follows. In section \ref{Related Work} we outline related works that motivate our analysis technique and past results on dataset analysis. In section \ref{Methods} we describe our methods. Section \ref{Experiments} contains our results on four standard image datasets, and section \ref{Discussion} provides additional discussion and potential implications of and extensions to our work. We conclude in section \ref{Conclusion}.

\section{Related Work} \label{Related Work}

In this paper, we are interested in understanding the relative importance of image data in common datasets. This concept is related to two areas of research: active learning and data distribution analysis. In active learning, the goal is to adaptively subsample data to reduce the number of observations required to train a model. This sampling selects the most informative data (which can change over the course of training). Similarly, in analyzing the distribution of data, we are interested in understanding the similarity of datapoints and how the data is distributed. This concept is relevant for active learning, where the most informative examples can be better identified if a data distribution is known.

\subsection{Active Learning}
Active learning is a machine learning paradigm that uses an oracle to interactively query data on which to train a model. The oracle chooses data to help the model learn, allowing for faster convergence and less data usage overall. An excellent survey on the field can be found in \cite{settles2012active}.

This approach is useful in several contexts. For situations where unlabeled data is abundant but labeling is expensive, an active learning approach can help select a limited subset of the data that needs to be labeled to train a model adequately. Active learning can also help decrease training time by effectively decreasing dataset size. 

Empirical results have demonstrated success for active learning \cite{settles2012active}. However, there is evidence that it may not always work, and there are situations where random sampling performs better than some active learning algorithms \cite{schein2007active, guo2008discriminative}. These results suggest that active learning may be dataset, model, and application dependent, at least to a certain extent. 

One model where active learning has seen success is support vector machines (SVM). Here, active learning has been applied to such tasks as image and text classification \cite{tong2001support, li2004multilabel} among others. Intuitively, active learning makes sense in the SVM case, as in practice the decision boundaries in SVMs often depend on only a few points in a dataset. 
Recent work suggests that active learning may be possible in the deep neural network setting \cite{sener2018active, ducoffe2018adversarial, huang2018cost}.

Our analysis approach is based on ideas in active learning and takes additional inspiration from the notion of importance given to support vectors in an SVM. In particular, we take the individual images that induce the largest gradient magnitudes to be our 'support vectors' in the sense that they are the most critical to training. Given the SVM intuition, we do not actively query this data, but rather fix a large batch of data at the start.

\subsection{Data Distribution Analysis}
Typical machine learning datasets are published with several surface-level properties. In computer vision tasks, this may include such information as image labels, image sizes, and image acquisition method (e.g., synthetically generated, crowd sourced, or curated). Depending on task, the image labels may include, for example, one or multiple class names (e.g., for classification), one or multiple object bounding boxes (e.g., for object detection), or an annotated figure (e.g., for pose recognition). Some datasets also contain additional annotation levels. For example, ImageNet contains a semantic tree that describes relations between categories. Additional distribution-level properties like mean pixel value or pixel value variance are easily computed. However, these properties do not fully model relation of the dataset to the true data manifold they describe.

There has been some relatively recent work such as \cite{torralba2011unbiased}, which analyze inherent bias in datasets. This bias is in relation to the world-view of the true data, and it reflects various biases in a specific dataset (e.g., in an image dataset containing 'car' images, this may show up as the angles at which a car is photographed, or the typical model of car photographed).
Various other work support these results and offer solutions to debias the data like weighting points according to dataset or other image properties \cite{khosla2012undoing, sugiyama2009dataset}. Some datasets also have more nefarious forms of biases due to limited diversity of geographical origin \cite{shankar2017no} or due to the underlying data having undesirable bias as in the corpus of English text \cite{bolukbasi2016man}.

In this paper, we seek to extend knowledge on the intrinsic properties of datasets by analyzing the relative importance of data in large image datasets. In particular, we are interested in the diversity / redundancy of data. Previous methods to measure diversity look at basic image-level statistics \cite{deng2009imagenet} or various human-annotated qualities like amount of texture or distinctiveness \cite{russakovsky2013detecting}. We believe it will be informative to investigate diversity from a trained model's perspective. Given the almost universal practice of using convolutional neural networks, we choose these networks as the models we investigate.

\begin{figure}
    \begin{center}
        \includegraphics[width=0.9\linewidth]{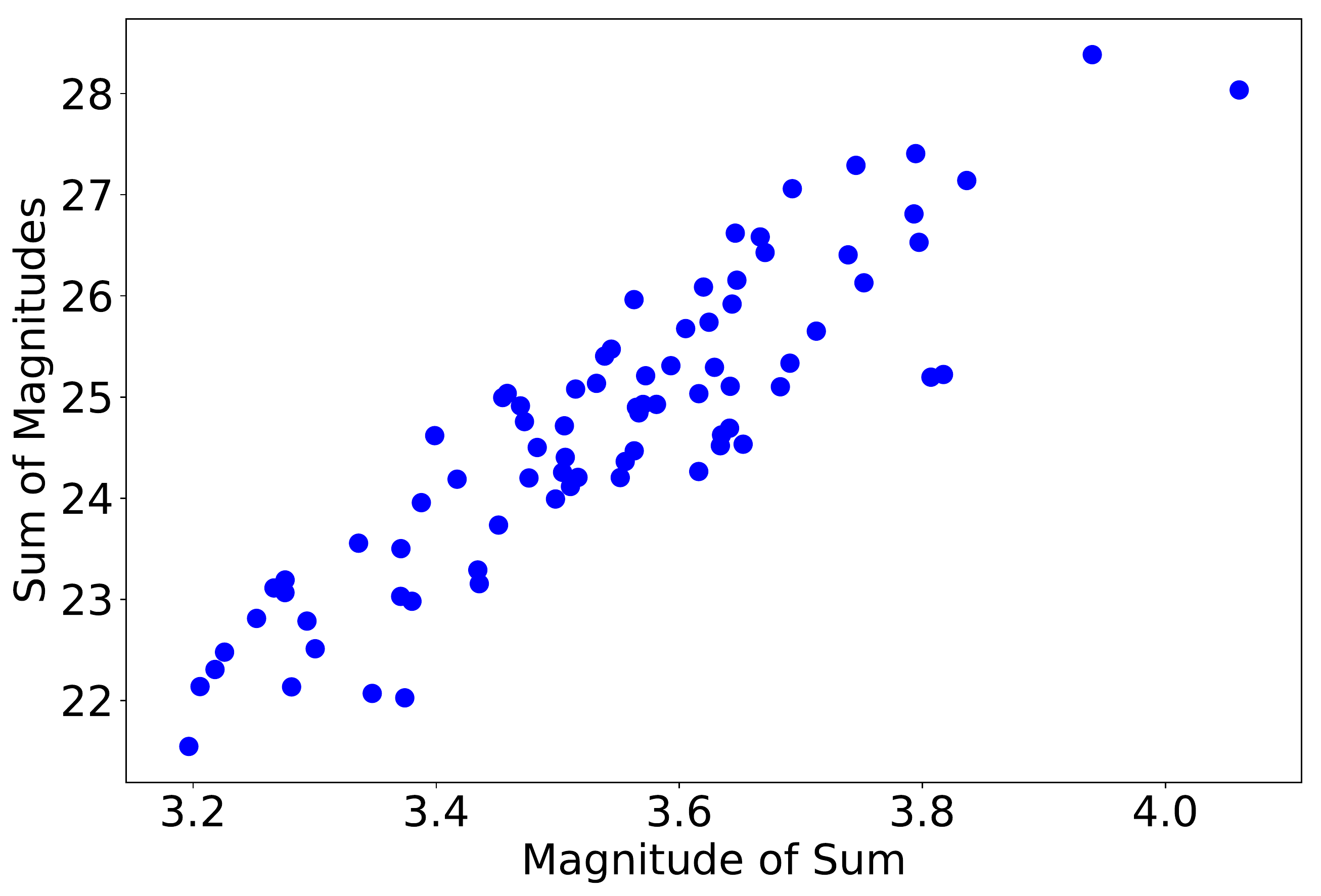}
    \end{center}
    \caption{Comparison of $||\sum_{i=1}^{N} \nabla_\theta L_{i,\theta}||$ and $\sum_{i=1}^{N} ||\nabla_\theta L_{i,\theta}||$ (see \eqref{eq:triangle-inequality}) for randomly sampled minibatches from ImageNet using the VGG16 model. Minibatch size is determined by the size used during training. Linear correlations suggest that the sum of gradient magnitudes is a reasonable approximation for ordering gradients.}
    \label{fig:grad_vis}
\end{figure}

 \begin{figure}[t]
    \begin{center}
        \includegraphics[width=0.9\linewidth]{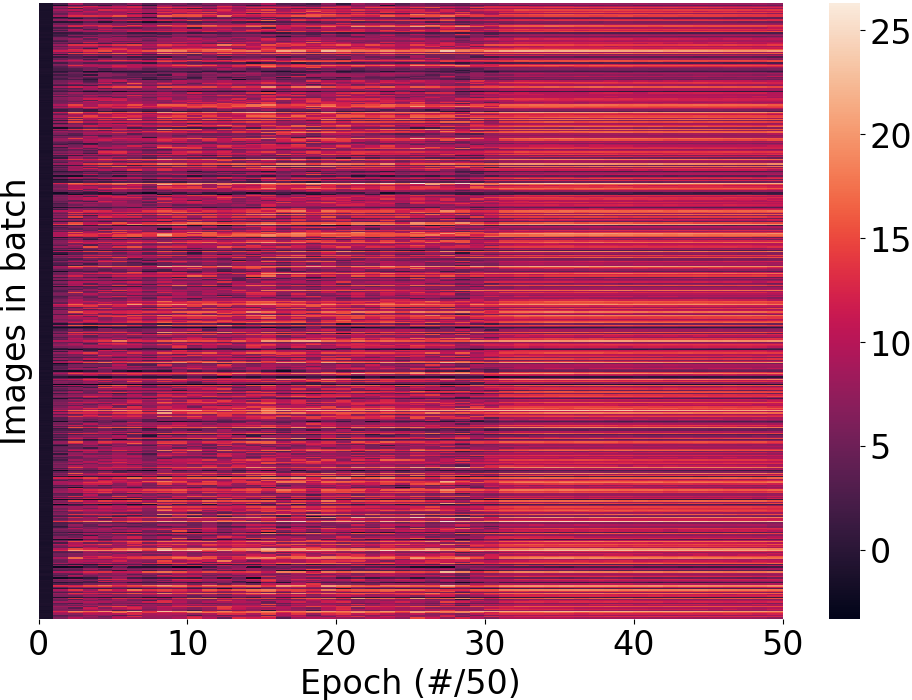}
    \end{center}
      \caption{Heatmap of gradient magnitudes for 512 MNIST images over 50 epochs of training. Black corresponds to large and white corresponds to small gradient magnitude. Color is on a log scale. Abrupt change at 30 occurs due to learning rate decay.}
    \label{fig:mnist-base-heatmap}
\end{figure}
    
 \begin{figure}[t]
    \begin{center}
        \includegraphics[width=0.9\linewidth]{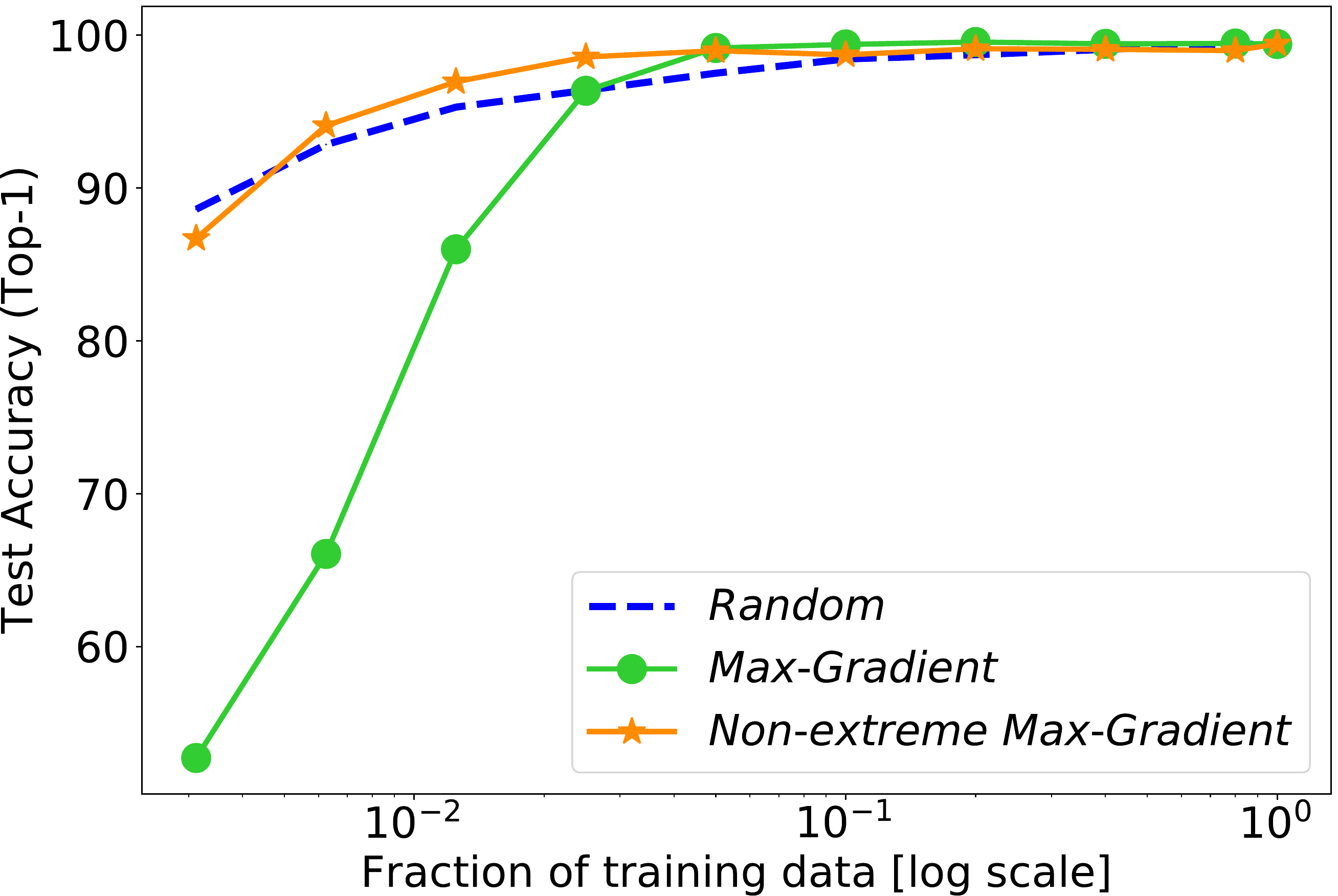}
   \end{center} 
       \caption{Top-1 test accuracy for MNIST. \textit{Non-extreme Max-Gradient} overtakes \textit{Random} when using 0.6\% of training data. \textit{Max-Gradient} overtakes \textit{Random} when using 3\% of training data.} %
    \label{fig:mnist-subsampledata4-accplot}
\end{figure}

\begin{figure*}
    \begin{center}
        \begin{subfigure}[t]{0.45\linewidth}
            \includegraphics[width=0.9\linewidth]{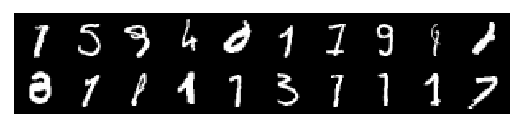}
            \caption{$99^{th}$ percentile}
        \end{subfigure}
        \begin{subfigure}[t]{0.45\linewidth}
            \includegraphics[width=0.9\linewidth]{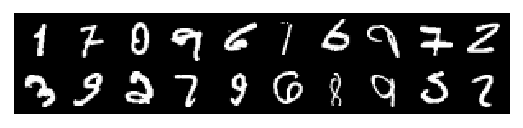}
            \caption{$95^{th}$ percentile}
        \end{subfigure}
        \begin{subfigure}[t]{0.45\linewidth}
            \includegraphics[width=0.9\linewidth]{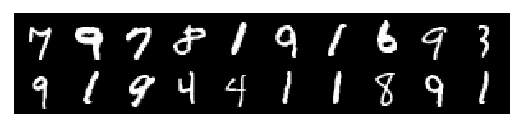}
            \caption{$50^{th}$ percentile}
        \end{subfigure}
        \begin{subfigure}[t]{0.45\linewidth}
            \includegraphics[width=0.9\linewidth]{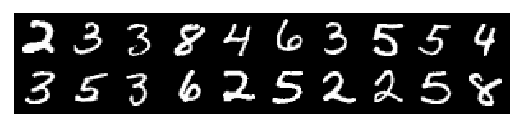}
            \caption{$5^{th}$ percentile}
        \end{subfigure}
   \end{center} 
       \caption{Sample images from MNIST organized by their final gradient magnitude.}
    
    \label{fig:mnist-images}
\end{figure*}

 \begin{figure*}
    \begin{center}
        \begin{subfigure}[t]{0.45\linewidth}
            \includegraphics[width=0.9\linewidth]{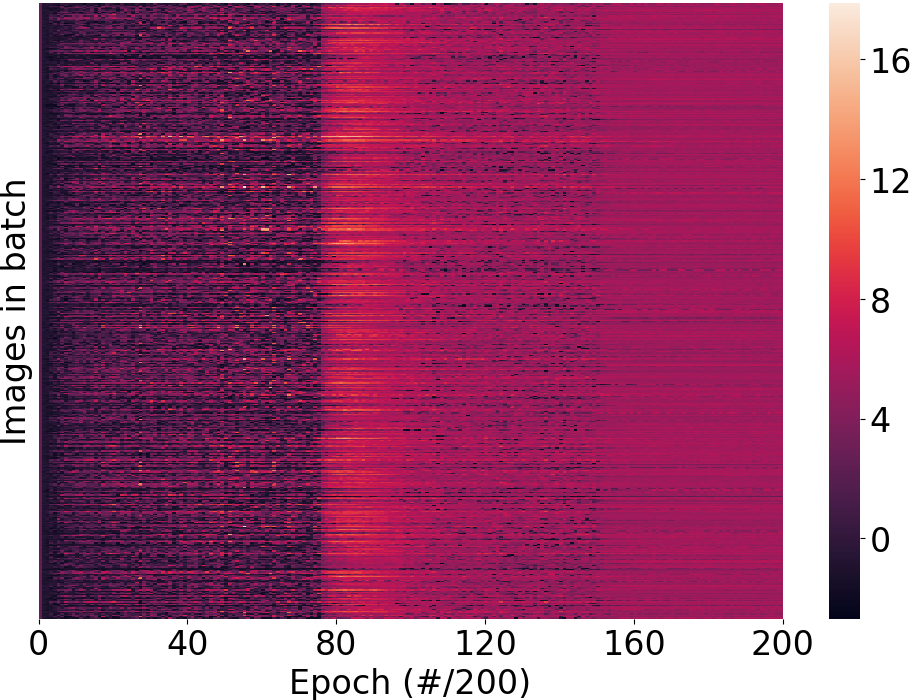}
            \caption{CIFAR-10 for VGG16 model; 200 epochs.}
            \label{fig:cifar10-vgg16-heatmap}
        \end{subfigure}
        \begin{subfigure}[t]{0.45\linewidth}
            \includegraphics[width=0.9\linewidth]{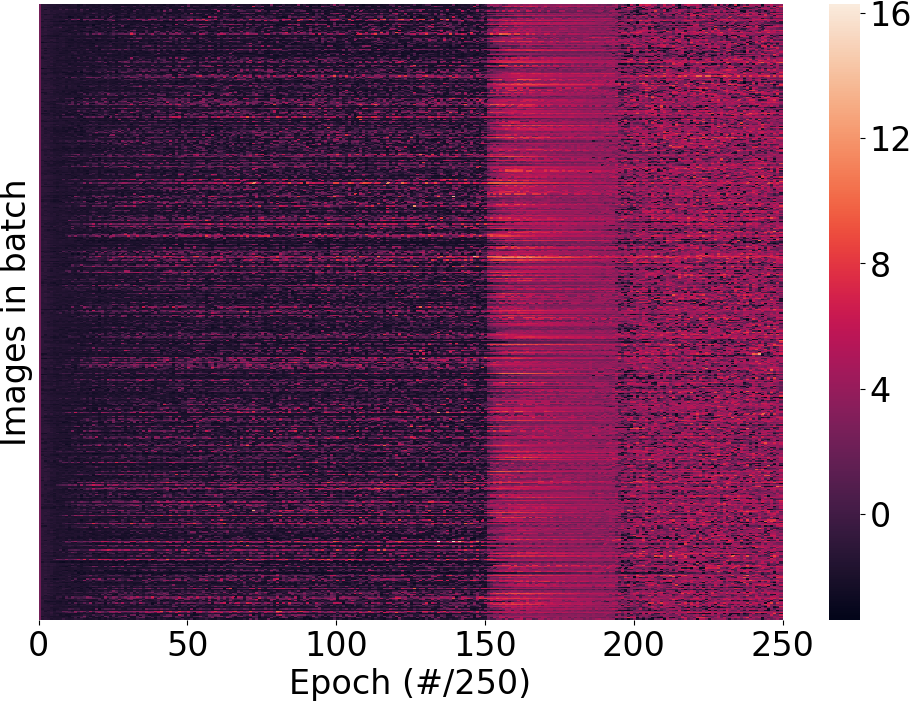}
            \caption{CIFAR-100 for VGG16 model; 250 epochs.}
            \label{fig:cifar100-vgg16-heatmap}
        \end{subfigure}
   \end{center} 
       \caption{Heatmap of gradient magnitudes for 512 CIFAR-10/100 images. Black corresponds to large and white corresponds to small gradient magnitude. Color is on a log scale.  Abrupt changes due to learning rate decay and likely saddle points.}
       \label{fig:cifar-vgg16-heatmap}
\end{figure*}

\begin{figure}[t]
    \begin{center}
        \includegraphics[width=0.9\linewidth]{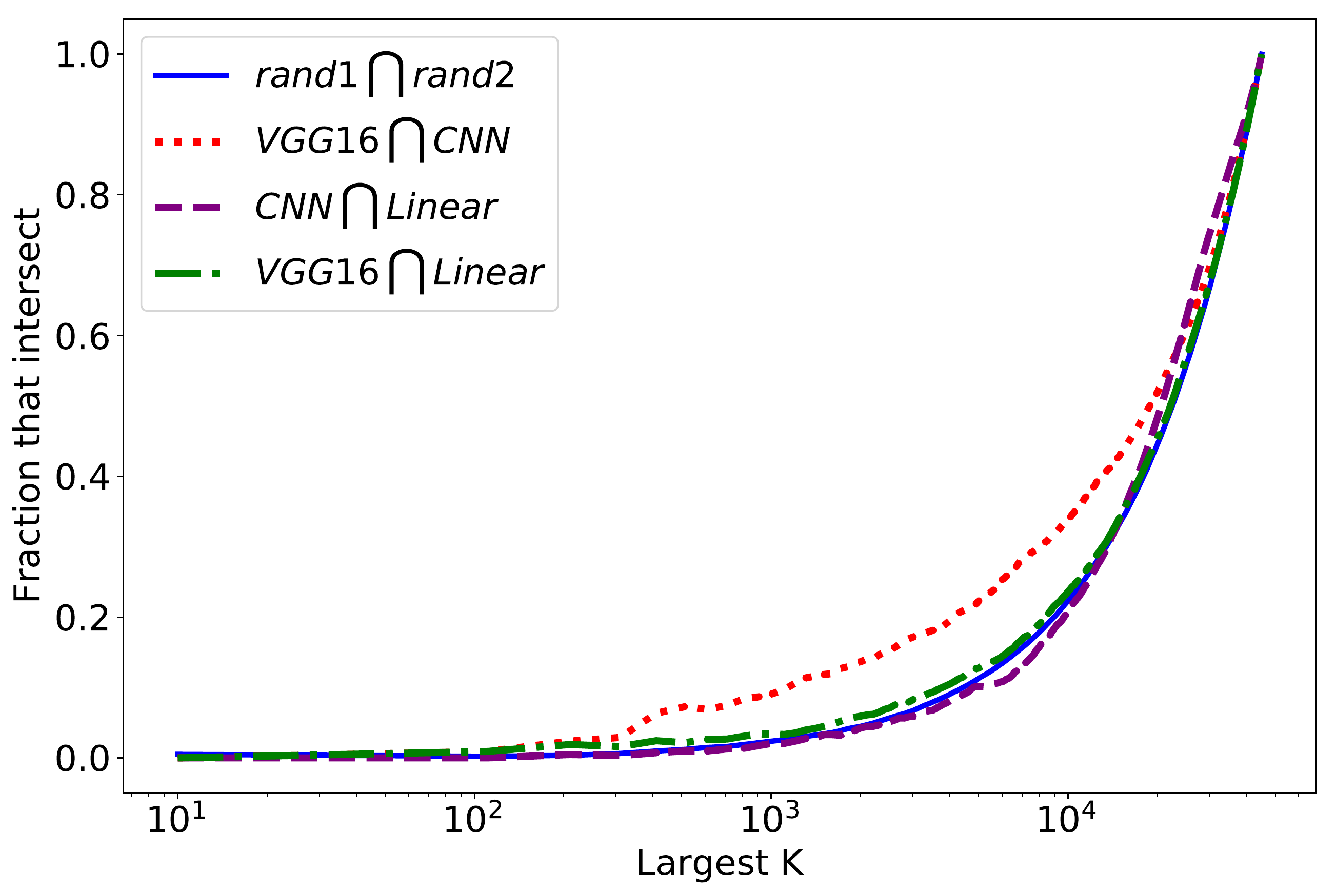} 
   \end{center} 
       \caption{This plot compares the similarity of the gradient-based ordering that different models impose on the CIFAR-10 training images. Let $S_{i,k}$ be the size-$k$ set of images with largest gradient magnitudes for model $i$. Then each plotted line computes $|S_{i_1,k} \bigcap S_{i_2,k}| / k$, as a function of $k$. $rand1 \bigcap rand2$ is a baseline that computes the size of intersection for two random size $k$ subsets. Interestingly, the size of the intersection between VGG16 and the generic CNN is significant (notice we plot with log-scaled X-axis).}
    
    \label{fig:cifar10-grad-comp}
\end{figure}

\begin{figure*}
    \begin{center}
        \begin{subfigure}[b]{0.45\linewidth}
            \includegraphics[width=0.9\linewidth]{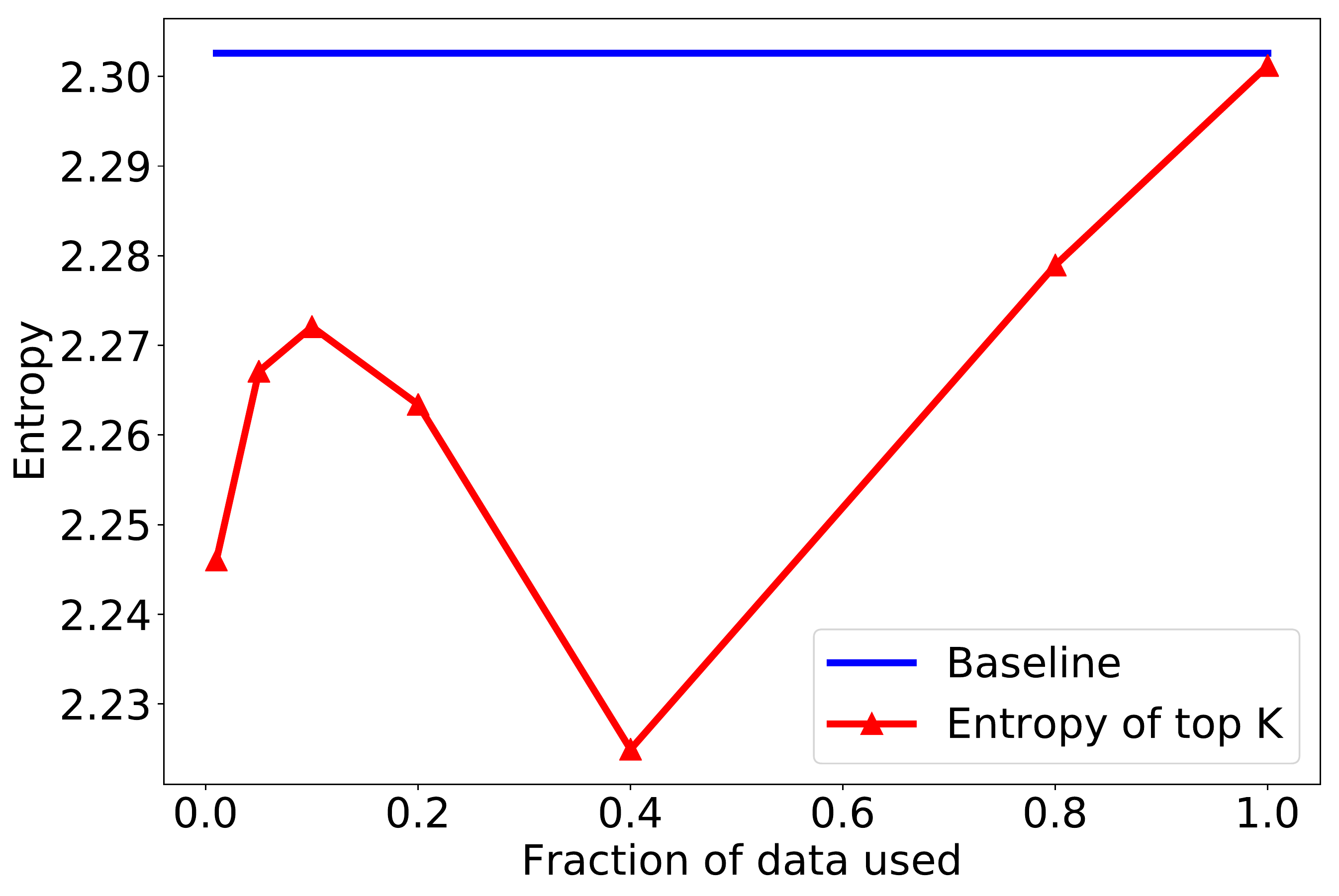}
            \caption{MNIST for (generic) CNN model}
        \end{subfigure}
        \begin{subfigure}[b]{0.45\linewidth}
            \includegraphics[width=0.9\linewidth]{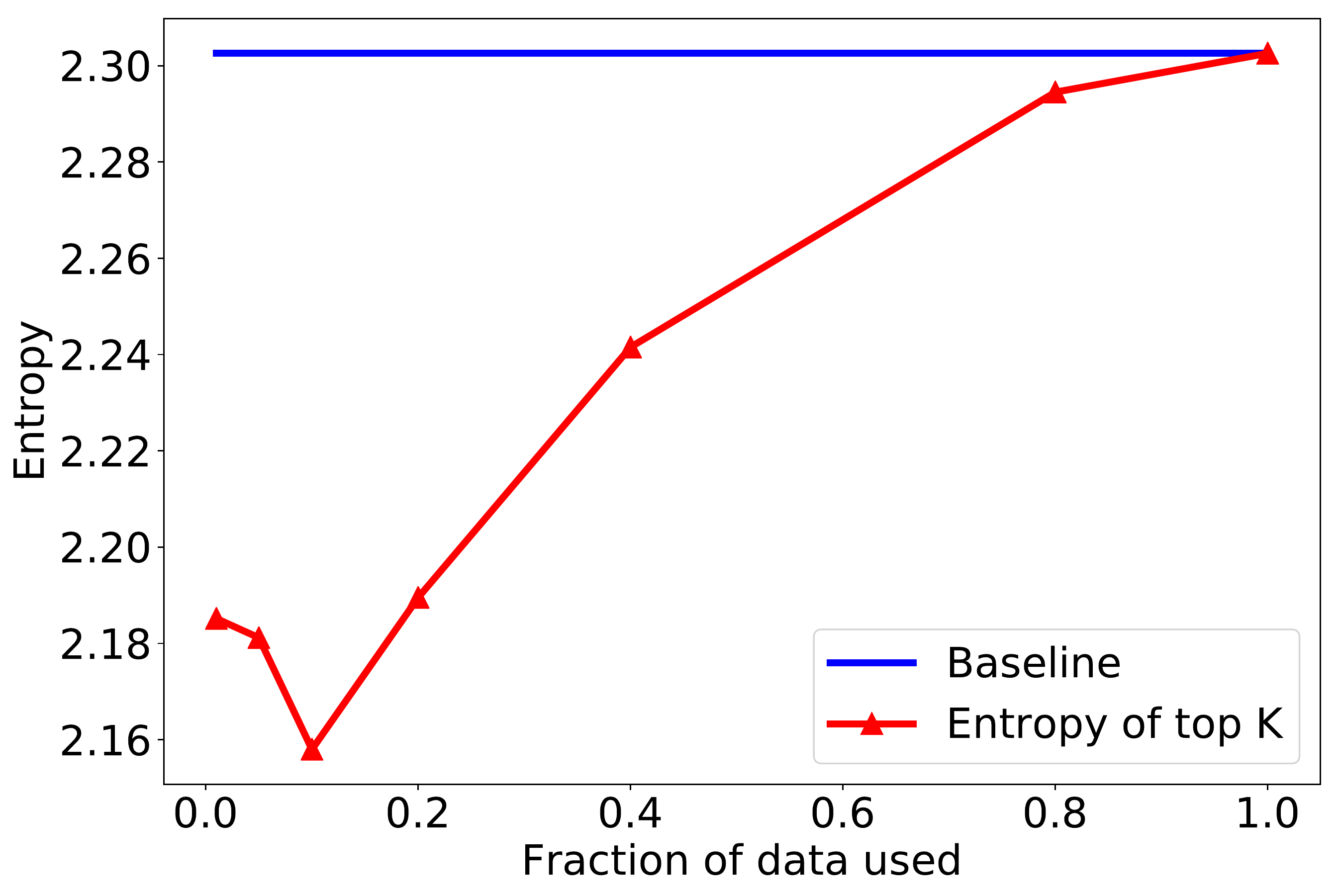}
            \caption{CIFAR-10 for VGG16 model}
        \end{subfigure}
        \begin{subfigure}[b]{0.45\linewidth}
            \includegraphics[width=0.9\linewidth]{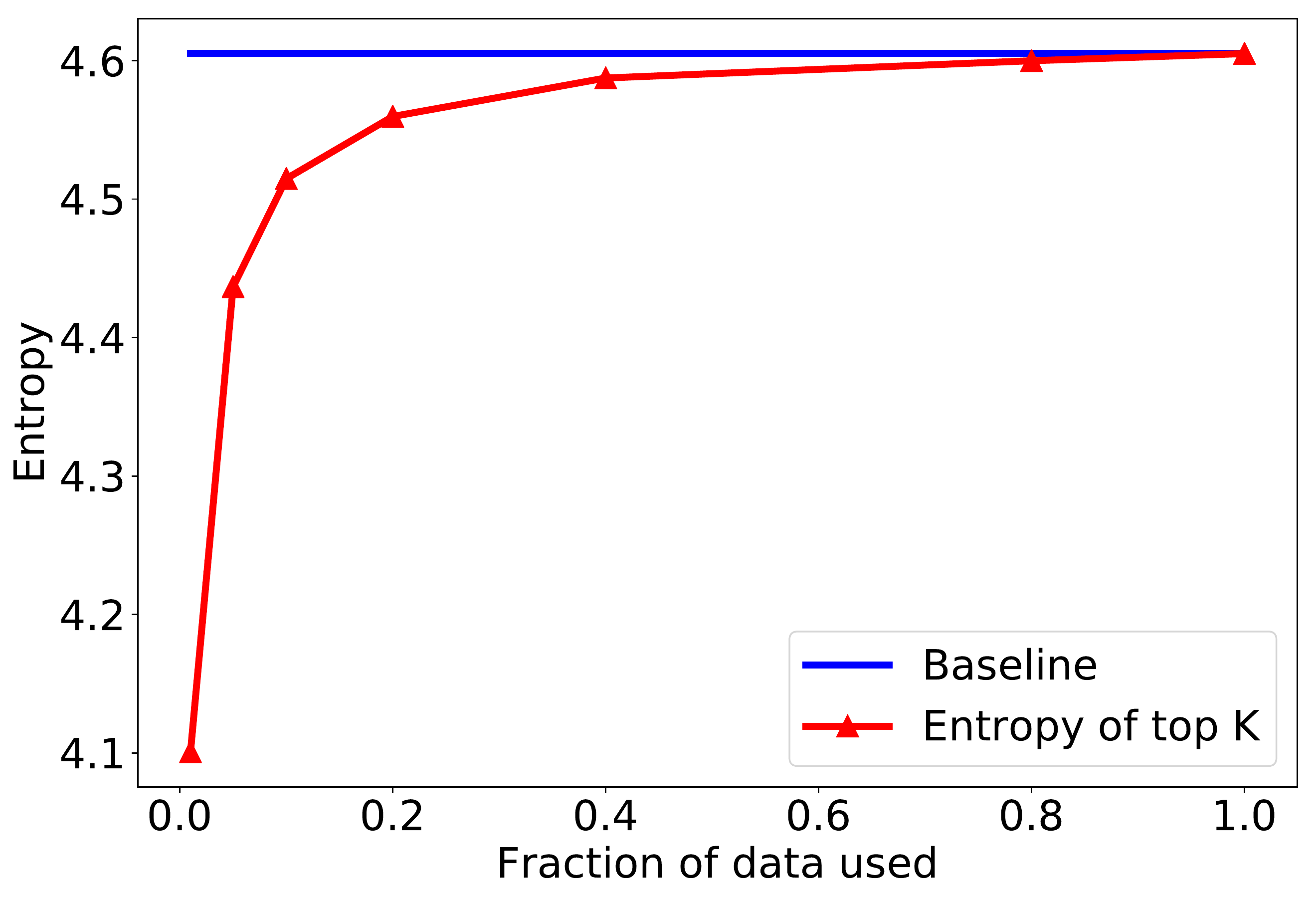}
            \caption{CIFAR-100 for VGG16 model}
        \end{subfigure}
        \begin{subfigure}[b]{0.45\linewidth}
            \includegraphics[width=0.9\linewidth]{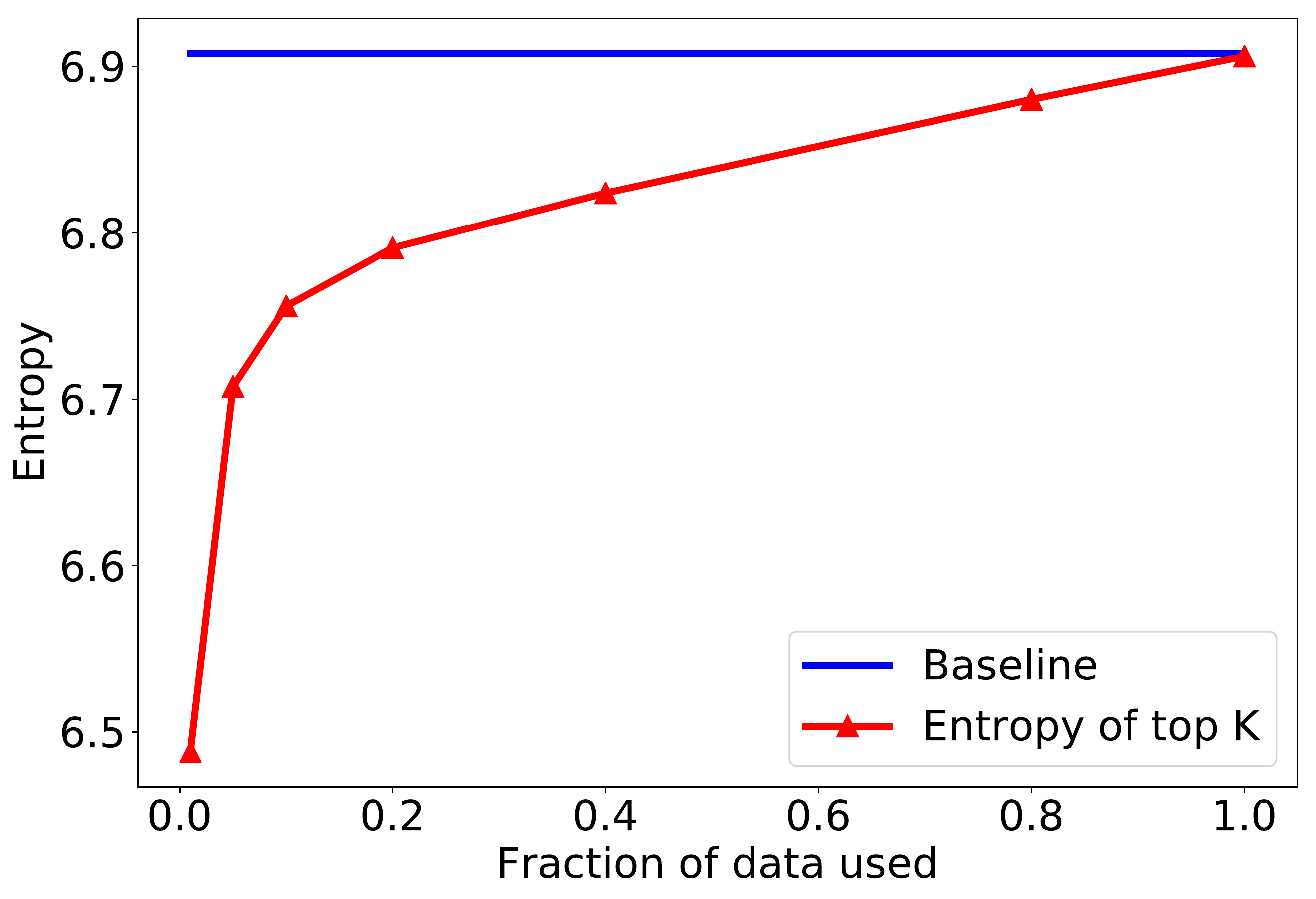}
            \caption{ImageNet for VGG16 model}
        \end{subfigure}
   \end{center} 
       \caption{Entropy of class labels for top-k images ordered by gradient magnitude on the given dataset and trained model. Baseline refers to the entropy for uniformly balanced label frequencies.}
    
    \label{fig:all-entropy}
\end{figure*}

 \begin{figure*}
    \begin{center}
        \begin{subfigure}[b]{0.45\linewidth}
            \includegraphics[width=0.9\linewidth]{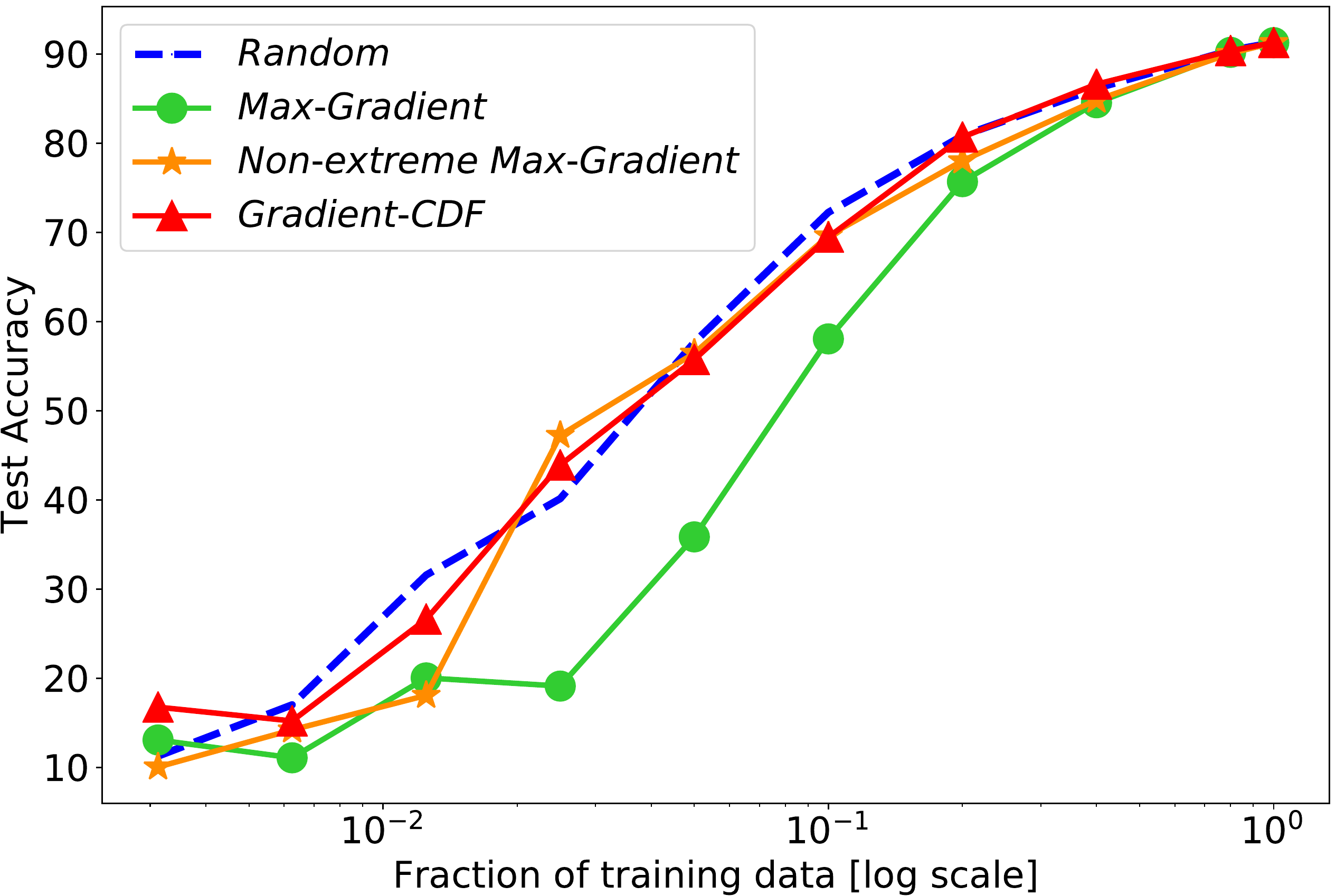}
            \caption{CIFAR-10 for VGG16 model}
            \label{fig:cifar10-vgg16-accplot}
        \end{subfigure}
        \begin{subfigure}[b]{0.45\linewidth}
            \includegraphics[width=0.9\linewidth]{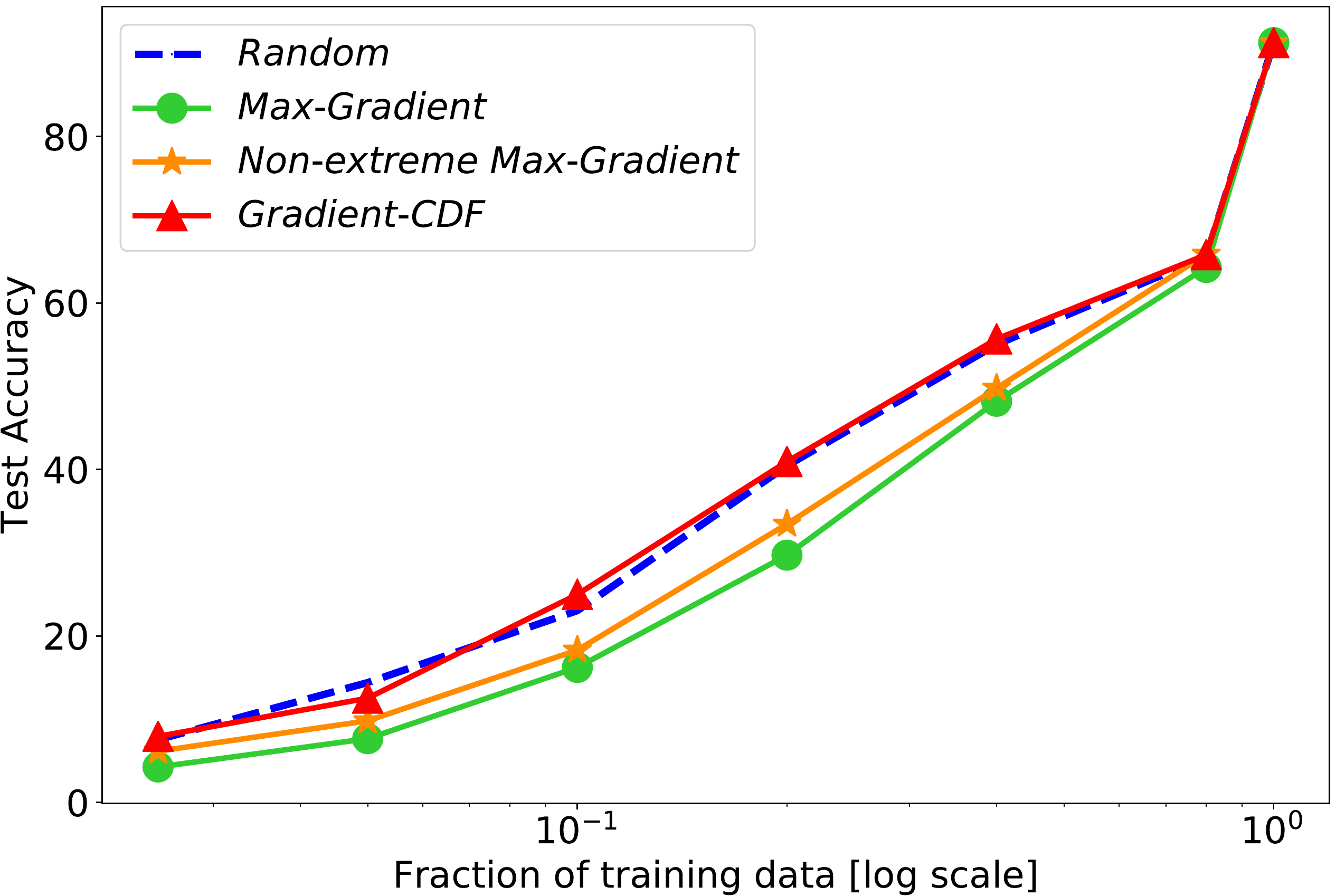}
            \caption{CIFAR-100 for VGG16 model}
            \label{fig:cifar100-vgg16-accplot}
        \end{subfigure}
   \end{center} 
       \caption{Top-1 test accuracy for CIFAR-10/100 run on a VGG16 network.}

\end{figure*}

 \begin{figure*}
    \begin{center}
        \begin{subfigure}[t]{0.45\linewidth}
            \includegraphics[width=0.9\linewidth]{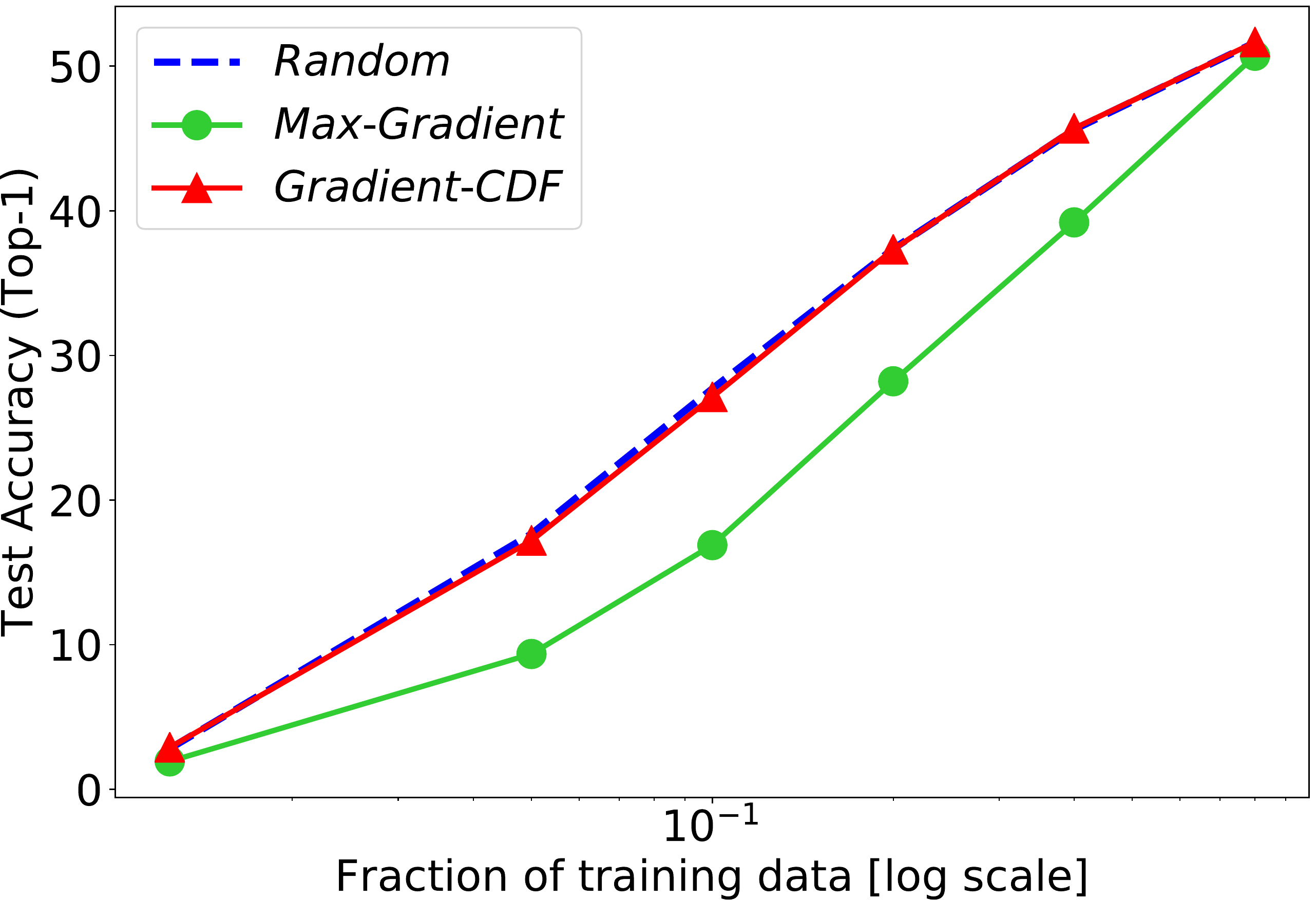}
            \caption{AlexNet; last data point uses 80\% of the dataset}
            \label{fig:imagenet-alexnet-accplot}
        \end{subfigure}
        \begin{subfigure}[t]{0.45\linewidth}
            \includegraphics[width=0.9\linewidth]{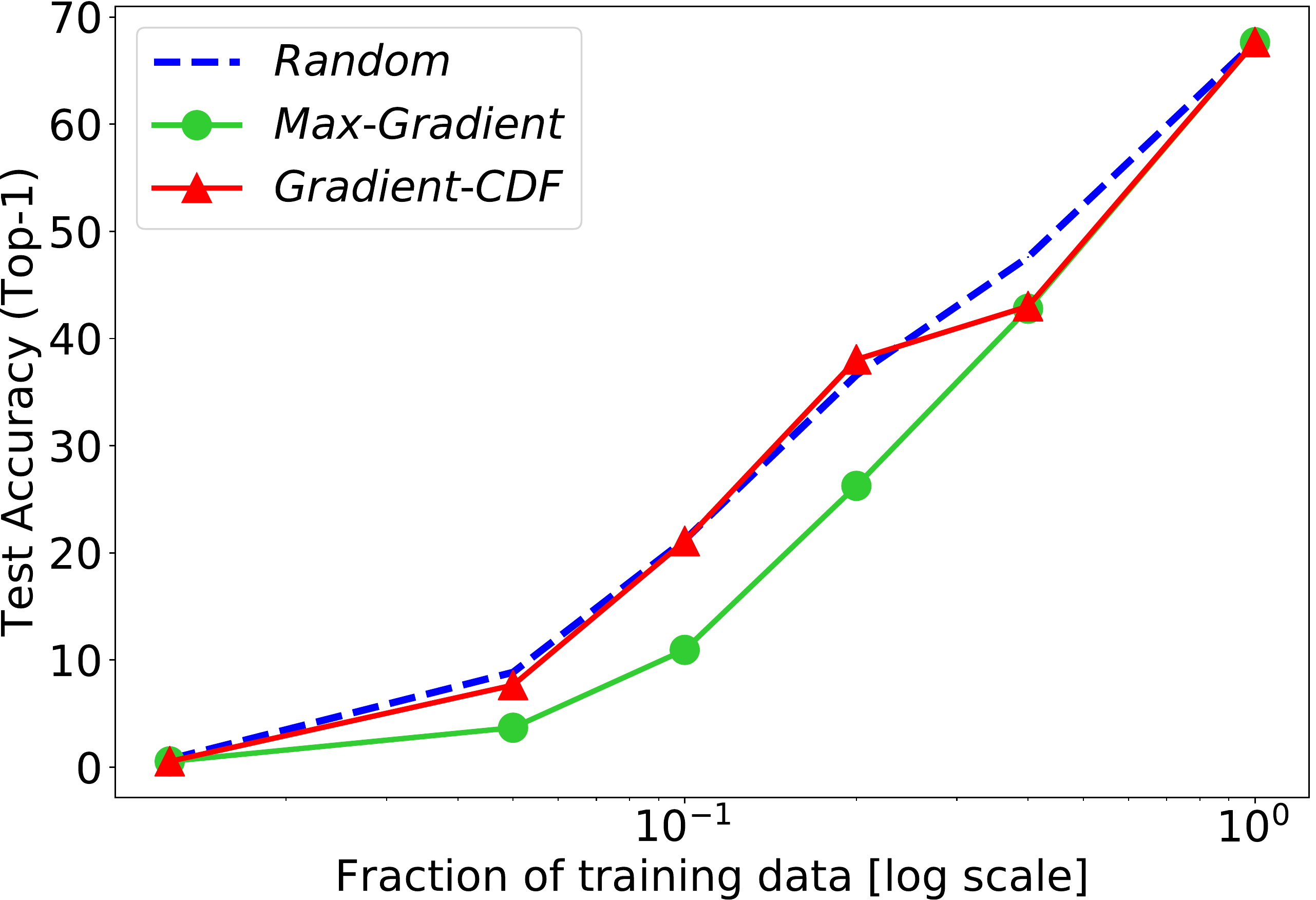}
            \caption{VGG16; last data point computed with pretrained model}
            \label{fig:imagenet-vgg16-accplot}
        \end{subfigure}
   \end{center} 
       \caption{Top-1 test accuracy for ImageNet. Gradients used from sampling computed using a VGG16 network.}
    \label{fig:imagenet-accplots}
\end{figure*}

\section{Methods} \label{Methods}

In Section \ref{Experiments}, we analyze four standard image datasets: MNIST, CIFAR-10, CIFAR-100, and ImageNet. For each dataset we select between one and three standard network architectures to perform our analysis. We also randomly subsample 10\% of the data for validation data prior to any analysis. Here we describe our analysis method.

\subsection{Gradient Magnitude as Importance Measure}
In our analysis, we need a measure of importance in order to subsample our set of training images. For this purpose, we will use the gradients of terms in the loss corresponding to individual training images w.r.t. the parameters of a fully trained deep network. Here we justify why this approach is reasonable.

In most cases, the training objective can be written in the following form:
\begin{align} 
    f^*_{\theta} &= \arg\min_{f_{\theta} \in \mathcal{F_{\theta}}} \mathcal{L}(f_\theta) \label{eq:opt}\\
    \text{where} \nonumber \\
    \mathcal{L}(f_\theta) &=  \left(\frac{1}{N}\sum_{i=1}^{N} l(f_{\theta}(\mathbf{x_i}), y_i)\right) + \mathcal{R}(f_{\theta})\\
        &= \left(\frac{1}{N}\sum_{i=1}^{N} L_{i,\theta}\right) + \mathcal{R}(f_{\theta}).
\end{align}
Here, $\mathbf{x_i}$ is the $i^{th}$ image and $y_i$ is its corresponding label, $N$ is the total number of images in the training set, $l$ is defined as the cross-entropy loss, $\mathcal{R}$ is some form of weight regularization (we omit the associated hyperparameters inside the function for clarity), and $f_{\theta}$ is our neural network (interpret as $\mathcal{F_\theta}$ is the set of possible neural networks given our standard architecture, and $\theta$ defines a specific set of parameters over that architecture).

So, the loss is a sum of losses over the training images. As the gradient is a linear operator, the gradient of the loss will be the sum of gradients over individual images. As is standard, we use a variant of stochastic gradient descent (SGD) to perform this optimization. Letting $\theta_t$ be our provisional network parameters after $t$ training iterations, $\eta_t$ be our learning rate at iteration $t$, and $\mathcal{B}_t$ be our minibatch at iteration $t$, we have
\begin{equation} \label{eq:grad-update}
\begin{split}
    \theta_{t+1} \leftarrow \theta_t &- \left(\frac{\eta_t}{|\mathcal{B}_t|}  \sum_{i \in \mathcal{B}_t} \nabla_\theta L_{i,\theta_t}\right) - \eta_t \nabla_\theta \mathcal{R}(f_{\theta_t})
\end{split}
\end{equation}

We consider the magnitude of change in parameters from one iteration to the next:
\begin{equation} \label{eq:grad-norm}
\begin{split}
    \left\Vert \theta_{t+1} - \theta_t  \right\Vert =  \left\Vert \left(\frac{\eta_t}{|\mathcal{B}_t|}  \sum_{i \in \mathcal{B}_t} \nabla_\theta L_{i,\theta_t}\right) - \eta_t \nabla_\theta \mathcal{R}(f_{\theta_t}) \right\Vert
\end{split}
\end{equation}

The magnitude of change in the parameters is directly related to how important/informative the current batch of training data is. If the current batch is important, then the model should change significantly after seeing the current batch. On the other hand, if it is not important, the model should remain almost the same. 

This observation doesn't directly tell us which individual training example is important. To find the important training examples, we upper bound the magnitude of change in the parameters. This is a conservative estimate of the importance of the batch, in the sense that a batch that is important is guaranteed to achieve a high upper bound, though there could be false positives, where a batch that is not important also achieves a high upper bound. This upper bound is obtained by applying the triangle inequality:

\begin{align} \label{eq:triangle-inequality}
    \left\Vert\sum_{i \in \mathcal{B}_t} \nabla_\theta L_{i,\theta}\right\Vert &\leq \sum_{i \in \mathcal{B}_t} \left\Vert\nabla_\theta L_{i,\theta}\right\Vert.
\end{align}

We can then find which training example contributes the most to this upper bound by selecting the examples that have the largest gradient magnitudes $\left\Vert\nabla_\theta L_{i,\theta}\right\Vert$. One possible concern is that some training examples that are selected may contribute a lot to the upper bound, but may not contribute significantly to the original quantity. However, this is justifiable because it is better to err on the conservative side and detect all examples that \emph{could} be important and cause large changes to our parameter values. One mitigating factor is that our models are highly expressive and can have millions of parameters. Because two gradient vectors are likely to be nearly orthogonal in high dimensions, the bound is unlikely to be very loose, since that would require the gradients of some training examples to be nearly colinear.

To confirm this intuition empirically, in Figure \ref{fig:grad_vis}, we plot randomly sampled minibatches from ImageNet showing the original quantity, $||\sum_{i=1}^{N} \nabla_\theta L_{i,\theta}||$ and the upper bound, $\sum_{i=1}^{N} ||\nabla_\theta L_{i,\theta}||$. The linear correlation between the quantities suggests that the upper bound is a good approximation for ordering gradient values, as larger individual gradient magnitudes tend to correspond to larger overall gradients for the batch.

Then we can select a size $k$ subsample, $\mathcal{B}$, based on this upper bound:
\begin{align}
    \mathcal{B}^* = \max_{\mathcal{B}:|\mathcal{B}|=k} \sum_{i=1}^{N} ||\nabla_\theta L_{i,\theta}||.
\end{align}
This set $\mathcal{B}^*$ is the set we choose by selecting the top-$k$ images with largest gradient magnitude.

\subsection{Subsample Selection} \label{Methods-Subsample}

    \begin{algorithm}
        \caption{Gradient Analysis} \label{alg:analysis}
        \begin{algorithmic}[1]
        \Procedure{analysis}{$f_{\theta}$}
            \State Train network $f_{\theta}$ on all data
            \State Compute test accuracy
            \For{$i=1,...,N$}
            \EndFor
            \State \Return \textproc{subsample\_analysis}$(f_{\theta})$
            \EndProcedure
        \end{algorithmic}
        
        \begin{algorithmic}[1]
        \Procedure{subsample\_analysis}{$f_{\theta}$}
            \State Subsample data using $\nabla_i$ \Comment{See Section \ref{Methods-Batch-Selection}}
            \State Retrain network $f_\theta$ on subsampled data only
            \State \Return data subsample, test accuracy
            \EndProcedure
        \end{algorithmic}
    \end{algorithm}
    
     Our analysis procedure for each dataset is described in Algorithm \ref{alg:analysis}. The main procedure, \textproc{Analysis}, can be broken into three steps:
    
    \begin{enumerate}
        \item [(2-3)] Train the network using the entirety of the training data, using validation data for early stopping. Log the test accuracy.
        \item [(4-5)] Compute the gradient of each network parameter with respect to a loss for each training image in the training dataset. We will use these gradients to subsample data in the next step.

        \item [(6)] Now retrain the network from a random initialization using a subsampled portion of the data. Then log the test accuracy as a measure of how well the subset represents the entirety of the dataset. We further describe how we subsample data in Section \ref{Methods-Batch-Selection}.

    \end{enumerate}

\subsection{Batch Selection} \label{Methods-Batch-Selection}

We propose three methods of sampling data based on gradients. We also include a random subsample baseline.

\begin{itemize}
    \item [a.] \textit{ Random}: This is our baseline approach. We randomly select the given number of images from all training images.
    \item [b.] \textit{Max-Gradient}: We select images in descending order by their gradient magnitude until we reach the given number of images.
    \item [c.] \textit{Non-extreme Max-Gradient}: We order images by their gradient magnitude in descending order. Then we discard the top 5\% of images, and proceed to select images in order until we reach the given number of images.
    \item [d.] \textit{Gradient-CDF}: Here, 'CDF' stands for 'cumulative distribution function.' We use the gradient magnitudes to induce a probability mass function (PMF) over the training images: Letting $g_i$ denote the gradient magnitude for the $i^{th}$ image, we define the PMF at $\mathbf{x_i}$ as
    \begin{equation}
        P(\mathbf{x_i}) = \frac{g_i}{\sum_{i=1}^{N} g_i}
    \end{equation}
    We subsequently use the resulting distribution to sample, without replacement, the given number of images.
\end{itemize}

\textit{Max-Gradient} is our original approach; the intuition is based in selecting images that have the largest affect on network weights through SGD. We find, however, that there are correlations between gradients that cause this method to artificially skew the distribution of images (see Figure \ref{fig:all-entropy}). 
Additionally, we observe that the images with largest gradients may just be outliers or poor examples (see Figure \ref{fig:mnist-images}). So we propose \textit{Non-extreme Max-Gradient} as an alternative that can decrease the number of outliers and increase diversity of images, and \textit{Gradient-CDF} as an alternative that softens how we select images by adding randomness into the process.

\section{Experiments} \label{Experiments}

We apply our analysis technique to four well-known image datasets: MNIST, CIFAR-10, CIFAR-100, and ImageNet. To compute the gradient magnitudes used in our subsampling procedure, there are several options. We considered the following norms to apply to the gradient vector: $\ell_1$, $\ell_2$, and $\ell_\infty$ norms. We also considered computing the norm over subsets of all parameters. In particular, (A) biases only, (B) weights only, and (C) last layer weights only (this may be reasonable as the magnitude of the gradient in each network layer can vary significantly). We observed little difference between these choices on the MNIST and CIFAR-10 datasets. Given computational limitations, we only report results using $\ell_2$ norm and option (C) for MNIST, CIFAR-10, and CIFAR-100; we use the $\ell_2$ norm and option (B) for ImageNet to provide some variety.

We consider the following standard networks: VGG16 \cite{simonyan2015very} and AlexNet \cite{krizhevsky2012imagenet}. We use a few additional shallow networks for comparisons on MNIST and CIFAR-10.

\subsection{MNIST} \label{Experiments-MNIST}
 
    For this study, we consider a network architecture that consists of 2 convolution layers followed by two fully connected layers. Given that classification on MNIST is generally considered easy, we did not tune this architecture. Indeed, this model attains 99.42\% test accuracy.
    .
    
    Before performing the main analysis, it is useful to understand the relative magnitudes of gradients at the end of training. In Figure \ref{fig:mnist-base-heatmap}, we plot a heatmap showing the negative log magnitude of gradients for 512 randomly sampled data points throughout training. At the beginning of training, all of these gradient magnitudes start roughly equal. Over time, all magnitudes decrease. By the end of training some images have much smaller gradients than others. So, effectively, towards the end of training, only a subset of training examples contribute significantly to the parameter updates. Intuitively, these correspond to the hard/more important training examples. 
    
    Now we present the result of our gradient analysis, subsampling data using several methods. In Figure \ref{fig:mnist-subsampledata4-accplot}, we plot the final top-1 test accuracy for \textit{Random}, \textit{Max-Gradient}, and \textit{Non-extreme Max-Gradient} subsampling. Notice that for very small batch sizes (corresponding to 175 total training images), random subsampling outperforms either gradient based approaches. Additionally, \textit{Non-extreme Max-Gradient} outperforms the regular \textit{Max-Gradient} approach. One possible explanation is that the top few images with largest gradients could be outliers and may not be representative of the dataset.

    However, when more data is used, \textit{Max-Gradient} outperforms the other two methods. This result is especially interesting as it suggests we need some `easy' images to be able to train the neural net successfully, but the hardest images (with the largest gradient magnitudes) are still the most important. As this leader change happens at only 5\% of the training data, and moreover because the test accuracy of \textit{Max-Gradient} is already nearly the test accuracy when using all the data, this suggests that MNIST has much redundant data and can be well approximated by a substantially condensed subset. The fact that this subset can be found by examining the magnitudes of gradients suggest that gradients are a reasonable measure of importance for each data example.

    It is also informative to visualize the training images with the largest gradient. Looking at Figure \ref{fig:mnist-images}, it is clear that images with largest gradient are difficult to classify, while the rest get progressively easier to read.

\subsection{CIFAR-10} \label{Experiments-CIFAR-10}

    We analyze CIFAR-10 using three network architectures to show how various models behave and to demonstrate how our analysis can apply in general. The architectures are: (1) a linear classifier, (2) and five layer CNN, and (3) VGG16 adapted for CIFAR-10.
    
    In Figure \ref{fig:cifar10-grad-comp}, we compare how the set of largest gradient images vary across network architectures. VGG16 and the CNN consistently have a higher overlap than the random baseline. However, this overlap is still a small percentage of the subsampled dataset size for small subsample amounts, suggesting that the images with largest gradient are somewhat specific to a given model. An interesting intuition here is that the most difficult images in a dataset can be model-specific rather than intrinsic to the image. 
    
    To determine whether gradients are correlated, we can compute the entropy of the class labels for the top-k images. In Figure \ref{fig:all-entropy}, we plot this entropy for different values of k. It can be seen that the entropy is lower when we select less data, confirming the existence of correlations between largest gradient and class label. Comparing the plots for MNIST and CIFAR-10, we see that there is greater skew in distribution for CIFAR-10, perhaps contributing to the decrease in test accuracy from \textit{Max-Gradient} subsampling. We take this as a motivation for using the \textit{Gradient-CDF} method.
    
    Now we look at some comparisons on the VGG16 models. In Figure \ref{fig:cifar10-vgg16-heatmap} we show how gradients for a randomly selected batch of training images change over training. The two abrupt changes result from learning rate decay. Also note that the gradients are colored on a log scale, so there are significant differences in gradient value at the end of training. However, compared to MNIST, the differences are less striking and suggest that the importance of images in CIFAR-10 is more distributed. Figure \ref{fig:cifar10-vgg16-accplot} compares the test accuracy of the final model trained using various subsampling procedures. We can see that  gradient-based sampling can give an increase in test accuracy, but random sampling can be better. This is in stark contrast to our results on MNIST where gradient-based sampling was clearly beneficial.
    
\subsection{CIFAR-100} \label{Experiments-CIFAR-100}

    CIFAR-100 is similar to CIFAR-10, but has 100 rather than 10 classes. We briefly report results of our analysis here. We use the VGG16 network adapted to the CIFAR-100 image size.

    Looking at the heatmap in Figure \ref{fig:cifar100-vgg16-heatmap}, it does seem like there are large variations in gradient magnitude at the end of training. However, the test accuracy plot in Figure \ref{fig:cifar100-vgg16-accplot} clearly suggests that gradient-based sampling performs worse than random sampling. Figure \ref{fig:all-entropy} shows that the largest gradients have a significantly skewed label distribution, potentially suggesting that certain classes are more difficult to classify than other classes.

\subsection{ImageNet} \label{Experiments-ImageNet}
    The ImageNet dataset consists of 1000 diverse classes and over a million images of varying sizes. As is standard, we scale all images down to 256x256 and take 224x224 croppings to train our network. We consider two network architectures: AlexNet and VGG16. For both architectures, we sample data using the gradient information from only the VGG16 architecture, giving us insight in how well the gradient-based importance measure generalizes between networks. Given the results from CIFAR-100 and CIFAR-10, we only include analysis for the following sampling methods: \textit{Random}, \textit{Max-Gradient}, and \textit{Gradient-CDF}.
    
    Because the test set of ImageNet is not publicly available, we use the official validation set as our test set (and, as is true for all our datasets, we subsample our train set to attain a train and validation set).

    For AlexNet, the results we see are similar to those from CIFAR-100. In particular, Figure \ref{fig:imagenet-alexnet-accplot} shows that the \textit{Max-Gradient} sampling method result in lower performance than either \textit{Random} or \textit{Gradient-CDF} sampling. For VGG16 we see very similar trends in Figure \ref{fig:imagenet-vgg16-accplot} until 40\% of data is sampled. This result is especially interesting as it suggests that the gradient-based importance values may be similar between these two deep networks.

\section{Discussion} \label{Discussion}

The results we have presented provide several insights. On MNIST, subsampling by maximum gradient gives higher performance, suggesting that there is indeed redundancy in the dataset. However, when we move to CIFAR-10 and CIFAR-100, random sampling performs better than \textit{Max-Gradient} and is most closely matched by \textit{Gradient-CDF}, which is simply weighted random sampling. One indicator for why this may be the case is in the heatmaps shown in Figure \ref{fig:cifar-vgg16-heatmap}. At the end of training, the gradient magnitudes are much closer together than they are in other datasets, perhaps suggesting that all images have roughly the same 'difficulty' as seen by the network. So, there may not be a small subset of data that captures the entirety of either CIFAR dataset; in other words, CIFAR seems to be diverse in the sense that it is not redundant.

We also note that gradient-based sampling may not always be optimal. Looking at Figure \ref{fig:all-entropy}, it is apparent that sampling by gradient skews the class distribution when we order by gradient magnitude, which in turn makes generalization more difficult.
Note that in CIFAR-100, \textit{Non-extreme Max-Gradient} results in lower test accuracy, while in CIFAR-10 it achieved roughly the same performance as \textit{Random}. This difference may be due to CIFAR-100 having 10 times as many classes, and so the issue of image distribution skew is exacerbated. 

We see the same behavior for ImageNet. In Figures \ref{fig:imagenet-vgg16-accplot} and \ref{fig:imagenet-alexnet-accplot}, \textit{Gradient-CDF} closely matches \textit{Random} in test accuracy, while \textit{Max-Gradient} achieves a significantly lower test accuracy. In Figure \ref{fig:all-entropy}, we see a similar shape as in CIFAR-100. However, interestingly, it appears that the top gradient magnitude images for ImageNet are more varied than for CIFAR-100. This may be due to an even greater diversity of images in ImageNet than in CIFAR-100.

These observations suggest two takeaways:
\begin{enumerate}
    \item [(1)] MNIST contains redundant data. CIFAR-10, CIFAR-100, and ImageNet contain little redundant data. This conclusion is especially interesting considering that the properties of CIFAR-10 and MNIST are similar in some ways: both contain the same number of classes, and number of bytes in CIFAR-10 images is only roughly 4x the number in MNIST images (i.e., the data size is on the same scale). There are two potential reasons: (1) the dataset itself is collected in such a way that the images are more diverse, and (2) the underlying space of dogs, cats, and other animals and vehicles is inherently larger than that for Arabic numerals, despite the image data representation size being relatively close for both datasets. While intuitively obvious to some degree, it is interesting to see explicit evidence supporting this hypothesis. 
    
    \item [(2)] It might be significantly more difficult to reduce the amount of training data while maintaining performance on CIFAR-10, CIFAR-100, and ImageNet than on MNIST, suggesting that active learning on CIFAR-10, CIFAR-100, and ImageNet might be substantially more challenging than on MNIST. 

\end{enumerate}

\section{Conclusion} \label{Conclusion}

Given the importance of large datasets in modern machine learning, it is critical to understand dataset properties in order to better exploit the data. In this paper, we have proposed an approach to empirically analyze the diversity of data and applied this method on four image datasets of varying complexity. For CIFAR-10, CIFAR-100, and ImageNet, we have found that most training examples are valuable, while for MNIST we found that most training examples are largely redundant. The results also suggest the large number of images in CIFAR-10, CIFAR-100, and ImageNet are indeed necessary, and it is difficult to reduce the amount of data while maintaining performance.

The results from our empirical analysis are specific to the deep learning models used in obtaining our results rather than a direct property of the dataset. Interestingly, however, we have found evidence that the analysis results do generalize between models suggesting that some images are important, independent of the learning model.

While we selected gradient magnitudes as a method for analyzing importance, it would be interesting in the future to look into other importance metrics. In particular, given that gradients are indeed correlated, it may be worthwhile to consider importance metrics that are evaluated on batches rather than individual images. This idea has been successfully utilized in the active learning setting \cite{guo2008discriminative, wang2015querying}. With this framework, it would be possible to enforce properties like approximate orthogonality of gradients inside a batch and high entropy of class labels, both of which are seemingly important for high performance on small datasets. However, there is a tradeoff in complexity with the possibility of exponential computational complexity (in the size of the subset). So, heuristics or approximate optimization approaches may be necessary.

{\small
\bibliographystyle{ieee}
\bibliography{egbib}
}

\title{Are All Training Examples Created Equal? An Empirical Study \\ \vspace{5pt} \large Supplementary Material}
\pagenumbering{gobble}
\maketitle

Here we present additional evidence that the examples with largest gradient magnitudes can be correlated. 
The following results further justify the use of the \textit{Gradient-CDF} method (see Section \ref{Methods-Batch-Selection}).

\begin{figure}[t]
    \begin{center}
        \begin{subfigure}[b]{0.9\linewidth}
            \includegraphics[width=0.9\linewidth]{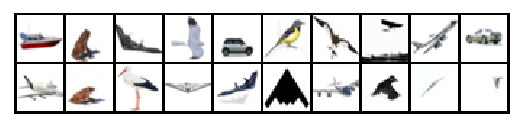}
            \caption{Linear model}
        \end{subfigure}
        \begin{subfigure}[b]{0.9\linewidth}
            \includegraphics[width=0.9\linewidth]{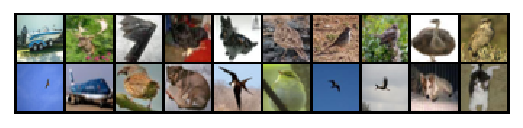}
            \caption{CNN model with vanilla architecture}
        \end{subfigure}
        \begin{subfigure}[b]{0.9\linewidth}
            \includegraphics[width=0.9\linewidth]{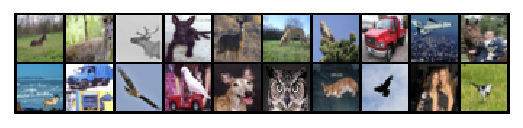}
            \caption{VGG16}
        \end{subfigure}
  \end{center} 
      \caption{Images from CIFAR-10 with largest gradient magnitudes, organized by network architecture. Note that these images share regularities: for the linear model, the images all have white background; for the vanilla CNN model, images of distant birds appear more commonly than usual. The presence of regularities in terms of pixel values (in the case of the linear model) or semantic categories (in the case of the vanilla CNN model) among images with high gradient magnitudes suggests that selecting images based purely on gradient magnitudes will result in correlated training examples. }

    \label{fig:cifar10-images000}
\end{figure}
\begin{figure}[t]
    \begin{center}
        \begin{subfigure}[b]{0.9\linewidth}
            \includegraphics[width=0.9\linewidth]{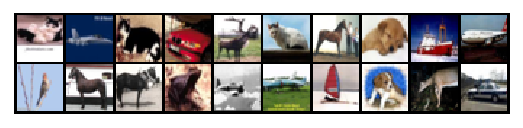}
            \caption{Linear model}
        \end{subfigure}
        \begin{subfigure}[b]{0.9\linewidth}
            \includegraphics[width=0.9\linewidth]{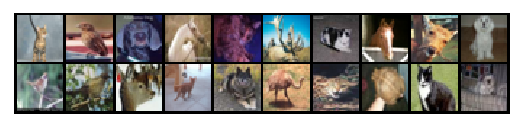}
            \caption{CNN model with vanilla architecture}
        \end{subfigure}
        \begin{subfigure}[b]{0.9\linewidth}
            \includegraphics[width=0.9\linewidth]{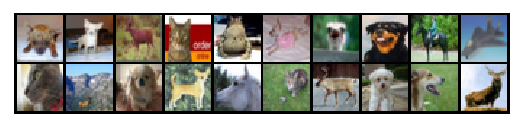}
            \caption{VGG16}
        \end{subfigure}

  \end{center} 
      \caption{Sample images from CIFAR-10 whose gradient magnitudes are in the top 5\%, organized by network architecture. Note the prevalence of cats and dogs in all models, suggesting category-level correlations.}
    
    \label{fig:cifar10-images005}
\end{figure}

\begin{figure*}
    \begin{center}
        \begin{subfigure}[b]{0.45\linewidth}
            \includegraphics[width=0.9\linewidth]{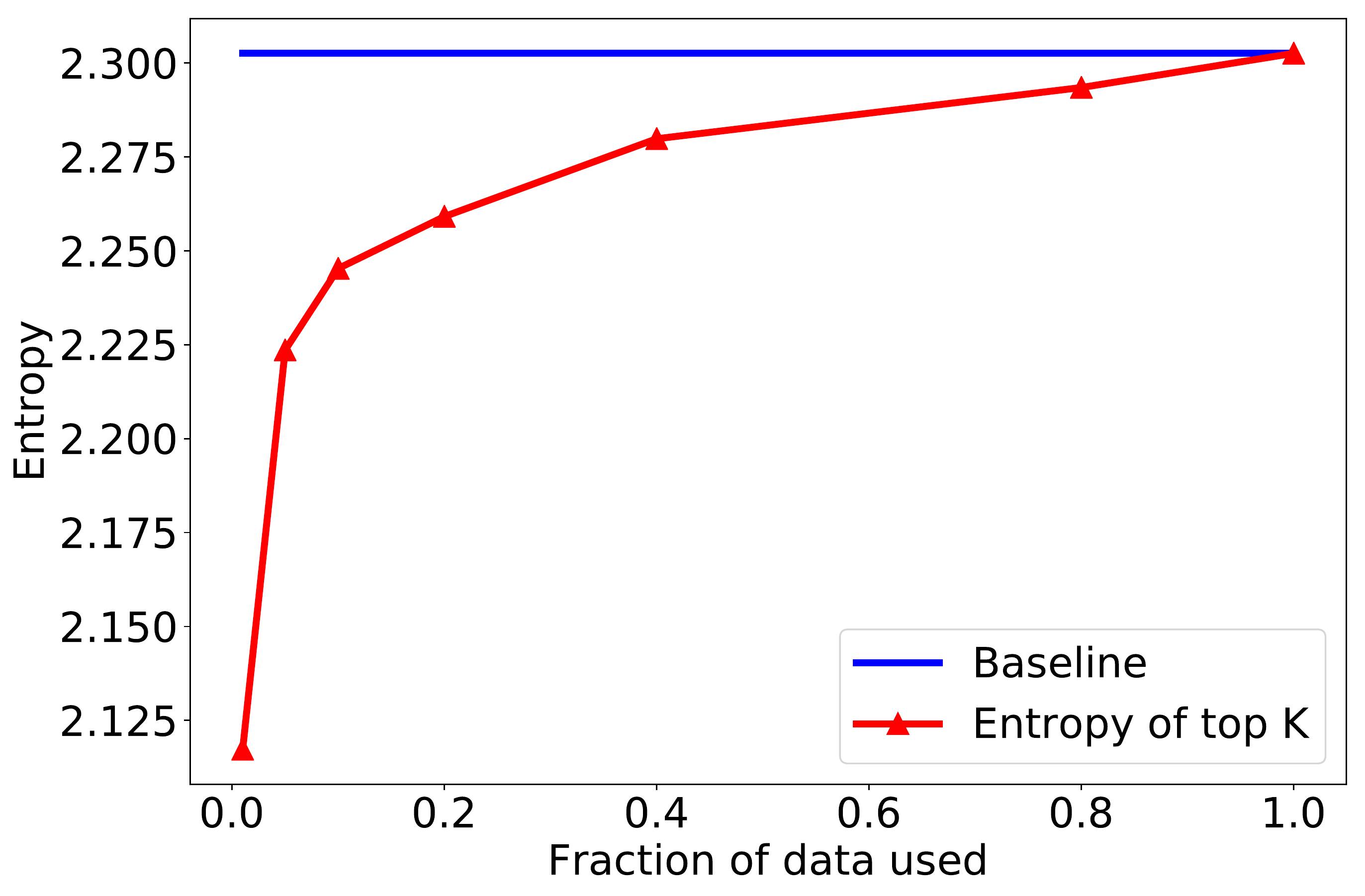}
            \caption{Linear model}
        \end{subfigure}
        \begin{subfigure}[b]{0.45\linewidth}
            \includegraphics[width=0.9\linewidth]{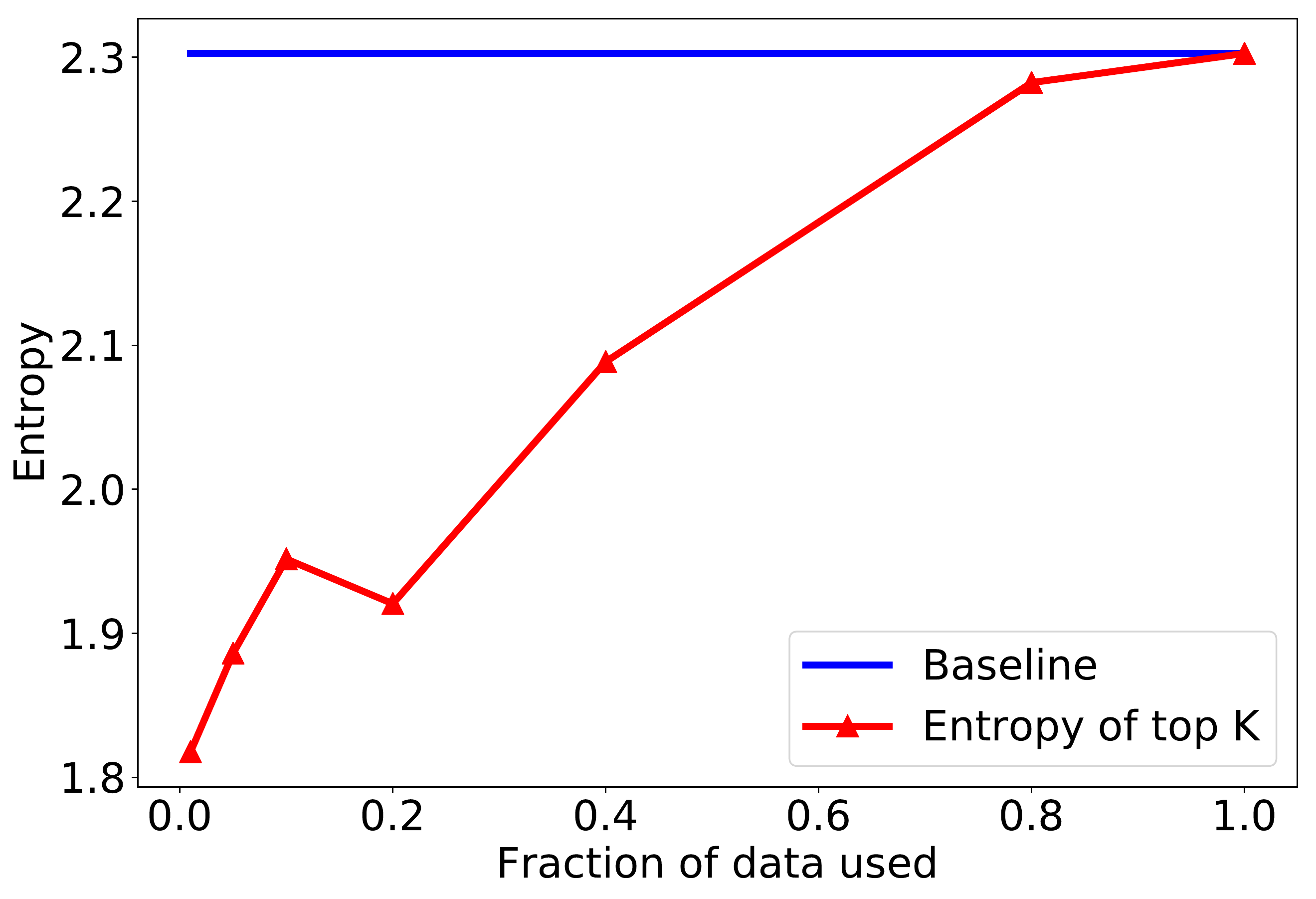}
            \caption{CNN model with vanilla architecture}
        \end{subfigure}
        \begin{subfigure}[b]{0.45\linewidth}
            \includegraphics[width=0.9\linewidth]{section_files/cifar10/vgg16/entropy.pdf}
            \caption{VGG16}
        \end{subfigure}
   \end{center} 
       \caption{Entropy over class labels for CIFAR-10 images in the top $x\%$ in terms of gradient magnitude for the given trained model. Baseline refers to the entropy for uniformly distributed labels.}
    
    \label{fig:cifar10-allentropy}
\end{figure*}

Gradients can be correlated at both the levels of semantic categories and pixels. For category-level correlations, certain categories tend to appear more frequently than others among images with high gradient magnitudes. For pixel-level correlations, certain colours tend to appear more frequently than others among images with high gradient magnitudes. 

In Figure \ref{fig:cifar10-images000} we display the images from the CIFAR-10 training set with largest gradient magnitudes in our three models (see Section \ref{Experiments-CIFAR-10}). The repeated images of birds in the the vanilla CNN model suggest class-based correlations, and the repeated white backgrounds for the linear model suggest pixel-level correlations. 

In Figure \ref{fig:cifar10-images005}, the prevalence of dog and cat images in both vanilla CNN and VGG16 models suggest class-based correlations. In Figure \ref{fig:cifar10-allentropy}, the low entropy for the CNN model at small subsamples of data indicates that a few classes of images have more images with large gradients than other classes and also suggests class-based correlations.

\end{document}